\documentclass[
]{ceurart}

\usepackage{listings}
\usepackage{svg}
\svgpath{{figures/}} 
\usepackage{graphicx} 
\usepackage{tikz}
\usetikzlibrary{positioning, shapes, arrows, fit, shadows, backgrounds, calc} 
\usepackage{caption}
\usepackage{placeins}
\usepackage[frozencache,cachedir=.]{minted}

\definecolor{customBlue}{HTML}{DFB2F4}
\definecolor{customRed}{HTML}{F49097}
\definecolor{customYellow}{HTML}{F5E960}
\definecolor{customGrey}{HTML}{55D6C2}
\definecolor{customLightBlue}{HTML}{F2F5FF}
\definecolor{backgroundcolor}{HTML}{FCFCFC}
\definecolor{verylightgray}{gray}{0.95}

\begin{document}

\copyrightyear{2024}
\copyrightclause{Copyright for this paper by its authors.
  Use permitted under Creative Commons License Attribution 4.0
  International (CC BY 4.0).}

\conference{3rd International Workshop on
Knowledge Graph Generation from Text (TEXT2KG)}

\title{KGValidator: A Framework for Automatic Validation of Knowledge Graph Construction}


 \author[1]{Jack Boylan}[%
 email=jackboylan@quantexa.com,
]
 \author[1]{Shashank Mangla}[%
 email=shashankmangla@quantexa.com,
 ]
 \author[1]{Dominic Thorn}[%
 email=dominicthorn@quantexa.com,
 ]
 \address[1]{Quantexa}
 \author[1]{Demian Gholipour Ghalandari}[%
 email=demiangholipour@quantexa.com,
 ]
 \author[1]{Parsa Ghaffari}[%
 email=parsaghaffari@quantexa.com,
 ]
 \author[1]{Chris Hokamp}[%
 email=chrishokamp@quantexa.com,
 ]


\begin{abstract}
This study explores the use of Large Language Models (LLMs) for automatic evaluation of knowledge graph (KG) completion models. Historically, validating information in KGs has been a challenging task, requiring large-scale human annotation at prohibitive cost. With the emergence of general-purpose generative AI and LLMs, it is now plausible that human-in-the-loop validation could be replaced by a generative agent. 
We introduce a framework for consistency and validation when using generative models to validate knowledge graphs. Our framework is based upon recent open-source developments for structural and semantic validation of LLM outputs, and upon flexible approaches to fact checking and verification, supported by the capacity to reference external knowledge sources of any kind. The design is easy to adapt and extend, and can be used to verify any kind of graph-structured data through a combination of model-intrinsic knowledge, user-supplied context, and agents capable of external knowledge retrieval.
\end{abstract}


\maketitle

\section{Introduction}
\label{sec:introduction}

Knowledge Graphs (KGs) are flexible data structures used to represent structured information about the world in diverse settings, including general knowledge \cite{10.1145/2629489wikidata}, medical domain models \cite{Kon2023SNOMEDCA}, words and lexical semantics  \cite{10.1145/219717.219748wordnet}, and semantics \cite{baker1998berkeley}. Most KGs are incomplete \citep{dong}, in the sense that there is relevant in-domain information that the graph does not contain. Motivated by this incompleteness, \textit{knowledge graph completion} research studies methods for augmenting KGs by predicting missing links \cite{chen2020knowledge}. 

\begin{figure}[ht!]
    \centering
    \resizebox{0.85\textwidth}{!}{%
        \begin{tikzpicture}[
    font=\ttfamily,
    square_node/.style args={#1#2}{
        rectangle, rounded corners, draw=none, fill=#1, 
        text width=#2, inner sep=2mm, align=left, font=\scriptsize,
    },
    square_node_header/.style args={#1#2#3#4}{
        rectangle, rounded corners, draw=none, fill=#1, 
        text width=#2, inner sep=2mm, align=left, font=\scriptsize,
        append after command={
            \pgfextra{
                \node [rectangle, draw, fill=white, text width=#3, align=center, font=\scriptsize, above=-0.5mm of \tikzlastnode.north] {#4};
            }
        },
    },  
    square_square_node/.style 2 args={
        rectangle, rounded corners, draw=none, fill=#1, 
        text width=#2, minimum height=#2, inner sep=1mm, align=center, font=\scriptsize,
    },    
    green_tick/.style={
        draw=none,
        thick,
        minimum size=0.5em,
        inner sep=0pt,
        path picture={ 
            \draw[green!70!black, thick, scale=0.4] (path picture bounding box.north east) -- (path picture bounding box.south) -- (path picture bounding box.west);
        }
    },
    red_cross/.style={
        draw=red,
        thick,
        cross out,
        minimum size=0.5em,
        inner sep=0pt,
    },
]
    \tikzstyle{arrow} = [very thick,->,>=stealth, draw=black!50]
    \tikzstyle{arrow_dashed_undirected} = [very thick, dashed, draw=black!50]
    \tikzstyle{arrow_dashed} = [very thick, dashed, ->, >=stealth, draw=black!50]
    

    \node[square_node={customLightBlue}{2.1cm}] (row1) {Ulysses};
    \node[square_node={customLightBlue}{2.1cm}, below=1mm of row1] (row2) {Moby Dick};
    \node[square_node={none}{2.1cm}, below=1mm of row2] (row3) {...};
    \node[square_node={customLightBlue}{2.1cm}, below=1mm of row3] (row4) {Finnegans Wake};
    \node[square_node={customLightBlue}{2.1cm}, below=1mm of row4] (row5) {Dubliners};
    \node[square_node={customLightBlue}{2.1cm}, below=1mm of row5] (row6) {Eveline};
    
    \node[green_tick, right=1.5cm of row1] (model1_pred1) {};
    \node[red_cross, right=1.5cm of row2] (model1_pred2) {};
    \node[red_cross, right=1.5cm of row4] (model1_pred3) {};
    \node[red_cross, right=1.5cm of row5] (model1_pred4) {};
    \node[green_tick, right=1.5cm of row6] (model1_pred5) {};

    \node[green_tick, right=1.5cm of model1_pred1] (model2_pred1) {};
    \node[red_cross, right=1.5cm of model1_pred2] (model2_pred2) {};
    \node[green_tick, right=1.5cm of model1_pred3] (model2_pred3) {};
    \node[green_tick, right=1.5cm of model1_pred4] (model2_pred4) {};
    \node[green_tick, right=1.5cm of model1_pred5] (model2_pred5) {};

    \node[square_square_node={red!50}{1.4cm}, above=0.5cm of model1_pred1] (testset) {Test Set};
    \node[square_square_node={green!40}{1.4cm}, above=0.5cm of model2_pred1] (testset) {External annotator / LLM};

    \node (data) [square_node_header={customLightBlue}{3.5cm}{2.2cm}{Incomplete triple}, left=3cm of row2] {
        (James Joyce, author of, \textcolor{red}{?})
    };

    \node (model) [square_node={customRed}{1.5cm}, below=1cm of data, xshift=1cm] {KGC model};    
    \node[square_node={none}{1.4cm}, left=5mm of row3] (predictions) {predictions};

    \draw [arrow] (data.south) to [out=-20,in=180, looseness=1.3] (model.west);
    \draw [arrow_dashed_undirected] (model.east) to [out=20,in=180, looseness=1.5] (predictions.west);
    \draw [arrow_dashed] (predictions.east) to (row3.west);

\end{tikzpicture}  
    }
    \caption{An example of the Closed-World Assumption in KG completion. Some of the triples predicted by a KG completion model are true in the real world (e.g. books written by James Joyce) but missing in the test set and would therefore be treated as false positives.}
    \label{fig:closed-world2}
\end{figure}
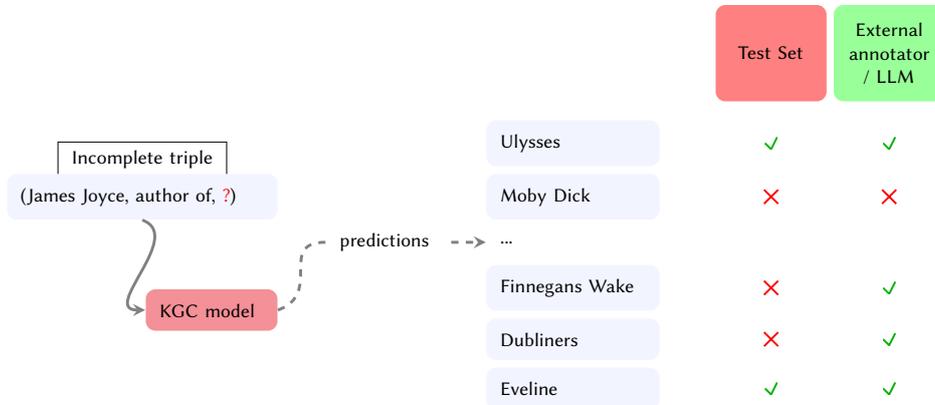

\paragraph{Challenges and Paradigms in KG Completion Evaluation:}
Evaluating KG completion models presents inherent challenges due to the natural incompleteness of most knowledge graphs (KGs) \cite{dong}. Traditional evaluation methods typically use a closed-world assumption (CWA), which deems absent facts to be incorrect, and may not effectively reflect the true capabilities of KG completion models \cite{closed-world-data-bases, lv-etal-2022-pre, sun-etal-2020-evaluation}. Alternatively, the open-world assumption (OWA) offers a more realistic framework by recognizing that KGs are inherently incomplete \cite{yang2022rethinking}. However, OWA complicates evaluation due to the need for extensive manual annotation of unknown triples, leading to significant time and cost implications. Efforts to improve the efficiency of human-driven KG evaluation include strategies like cluster sampling, which aims to reduce costs by modeling annotation efforts more economically \cite{gao2019efficient}. An illustration of these evaluation paradigms is shown in Figure \ref{fig:closed-world2}.

\begin{figure}[t!]
    \centering
    \resizebox{\textwidth}{!}{%
        \begin{tikzpicture}[
    font=\ttfamily,
    bigbox/.style args={#1#2#3#4}{
        rounded corners, fill=#1, draw=gray, align=center, inner sep=#2, line width=0.1pt,
        append after command={
            \pgfextra{
                \node [rectangle, draw, fill=white, text width=#3, align=center, font=\scriptsize, above=-0.5mm of \tikzlastnode.north] {#4};
            }
        }
    },
    square_node/.style 2 args={
        rectangle, rounded corners, draw=none, fill=#1, 
        text width=#2, minimum height=#2, inner sep=1mm, align=center, font=\scriptsize,
    },
    square_node_header/.style 2 args={
        rectangle, rounded corners, draw=none, fill=#1, 
        text width=4.7cm, inner sep=1mm, align=left, font=\scriptsize,
        append after command={
            \pgfextra{
                \node [rectangle, draw, fill=white, text width=2.5cm, align=center, font=\scriptsize, above=-1mm of \tikzlastnode.north] {#2};
            }
        }
    },
    neuron/.style={circle, draw=none, fill=customRed, minimum size=0.25cm},
    pics/llm/.style={
        code={
            \begin{scope}[local bounding box=llm]
              \foreach \i in {1,...,3}
                \node[neuron] (-input-\i) at (0,\i*0.5-1) {};
              \foreach \i in {1,...,4}
                \node[neuron] (-hidden-\i) at (1,\i*0.5-1.25) {};
              \node[neuron] (-output) at (1.75,0.05) {};
              \foreach \i in {1,...,3}
                \foreach \j in {1,...,4}
                  \draw (-input-\i) -- (-hidden-\j);
              \foreach \i in {1,...,4}
                \draw (-hidden-\i) -- (-output);
            \end{scope}
        }
    }
]
    \tikzstyle{arrow} = [very thick,->,>=stealth, draw=black!50]
    \tikzstyle{arrow_dashed} = [very thick, dashed, ->, >=stealth, draw=black!50]
    
    
    \node (web) [square_node={customBlue}{1.3cm}] {Web};
    \node (kg) [square_node={customBlue}{1.3cm}, right=0.25cm of web] {Wikidata};
    \node (docs) [square_node={customBlue}{1.3cm}, right=0.25cm of kg] {Docs};

    \pic [below=3cm of kg, xshift=-1cm] {llm};

    \begin{scope}[on background layer]
        \node (external_data_box) [bigbox={verylightgray}{2mm}{2cm}{External data}, fit=(web) (kg) (docs)] {};
    \end{scope};
    \begin{scope}[on background layer]
        \node (external_llm_box) [bigbox={verylightgray}{2mm}{1cm}{LLM}, fit=(llm)] {};
    \end{scope};

    \node (unvalidated_triple1) [square_node_header={customLightBlue}{Unvalidated triple}, left=2cm of external_data_box, yshift=-2.2cm] {
        \{\\
        \ \ ``subject": ``anaheim\_ducks",\\
        \ \ ``relation": ``teamplaysport",\\
        \ \ ``object": ``football"\\
        \}\\
    };
    \node (unvalidated_triple2) [square_node_header={customLightBlue}{Unvalidated triple}, below=1.5cm of unvalidated_triple1] {
        \{\\
        \ \ ``subject": ``alabama\_crimson\_tide",\\
        \ \ ``relation": ``teamplayssport",\\
        \ \ ``object": ``american football"\\
        \}
    };
    
    \node (validated_triple1) [square_node_header={customLightBlue}{Validated triple}, right=2cm of external_data_box, yshift=-1.4cm] {
        \{\\
        \ \ ``subject": ``anaheim\_ducks",\\
        \ \ ``relation": ``teamplaysport",\\
        \ \ ``object": ``football",\\
        \ \ \textcolor{red}{``is\_valid": False,}\\
        \ \ \textcolor{red}{``reason": ``The given context states that the Anaheim Ducks are actually an ice hockey team."}\\
        \}    
    };
    \node (validated_triple2) [square_node_header={customLightBlue}{Validated triple}, below=1.5cm of validated_triple1] {
        \{\\
        \ \ ``subject": ``alabama\_crimson\_tide",\\
        \ \ ``relation": ``teamplayssport",\\
        \ \ ``object": ``american football",\\
        \ \ \textcolor{green}{``is\_valid": True,}\\
        \ \ \textcolor{green}{``reason": ``The Alabama Crimson Tide represent the University of Alabama in the sport of American football."}\\
        \}    
    };

    \draw [arrow] (unvalidated_triple1.east) to [out=20,in=180] (external_llm_box.west);
    \draw [arrow] (unvalidated_triple2.east) to [out=-20,in=-180] (external_llm_box.west);
    \draw [arrow] (external_llm_box.east)  to [out=20,in=180]  (validated_triple1.west);
    \draw [arrow] (external_llm_box.east)  to [out=-20,in=180]  (validated_triple2.west);
    \draw [arrow_dashed] (web.south)  to [out=-20,in=180, looseness=1.5]  (external_llm_box.north west);
    \draw [arrow_dashed] (kg.south)  to [out=-20,in=180, looseness=1.5]  (external_llm_box.north west);
    \draw [arrow_dashed] (docs.south)  to [out=-20,in=180, looseness=1.5]  (external_llm_box.north west);

\end{tikzpicture}  
    }
    \caption{Framework for Validating Knowledge Graph Triples.}
    \label{fig:main-diagram}
\end{figure}
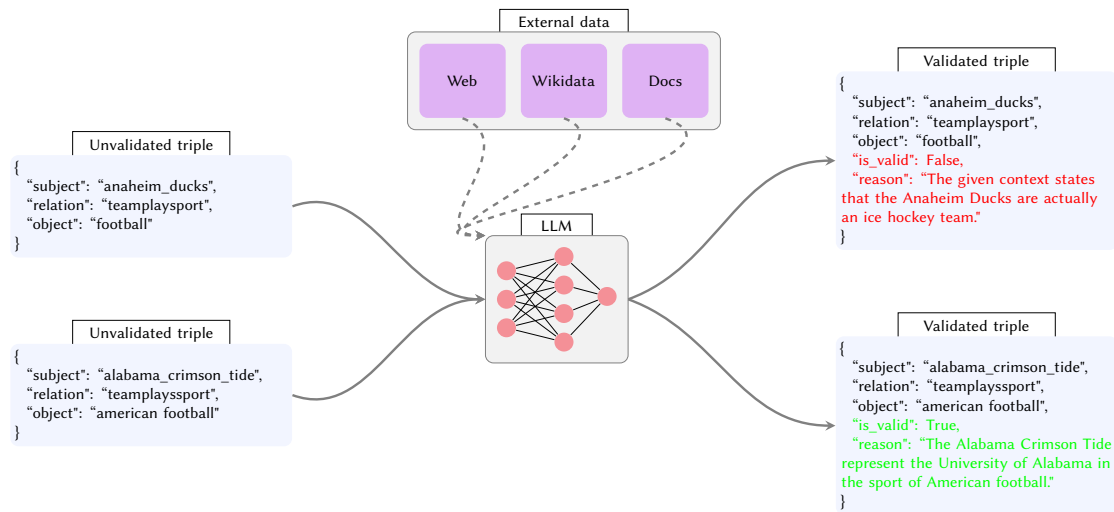

\paragraph{KGValidator Framework:}

Motivated by these challenges, we introduce \texttt{KGValidator} as a flexible framework to evaluate KG Completion using LLMs. At its core, this framework validates the triples that make up a KG using context. This context can be the inherent knowledge of the LLM itself, a collection of text documents provided by the user, or an external knowledge source such as Wikidata or an Internet search (refer to Figure \ref{fig:main-diagram} for a high-level overview). Importantly, our framework does not require any gold references, which are often only available for popular benchmark datasets. This enables evaluation of a wider range of KGs using the same framework.

\texttt{KGValidator} makes use of the Instructor\footnote{\url{https://github.com/jxnl/instructor}} library, Pydantic\footnote{\url{https://docs.pydantic.dev/}} classes, and function calling to control the generation of validation information. This ensures that the LLM follows the correct guidelines when evaluating properties, and outputs the correct data structures for calculating evaluation metrics. Our main contributions are:

\begin{itemize}
    \item A simple and extensible framework based on open-source libraries that can be used to validate KGs with the use of LLMs\footnote{Unfortunately, IP restrictions currently prevent us from sharing our implementation, but we are happy to directly correspond with interested researchers who wish to reproduce our results}.
    \item An evaluation of our framework against popular KG completion benchmark datasets to measure its effectiveness as a KG validator.
    \item An investigation of the impact of providing additional context to SoTA LLMs in order to augment evaluation capabilities.
    \item A straightforward protocol for implementing new validators using any KG alongside any set of knowledge sources.
\end{itemize}

\noindent The rest of the paper is structured as follows: Section \ref{sec:background} discusses key related work, Section \ref{sec:approach} covers our approach in detail, Section \ref{sec:experiments} presents several experiments designed to validate the framework, and Section \ref{sec:discussion} discusses results and possible extensions to this work.

\section{Background}
\label{sec:background}

\subsection{Knowledge Graph Construction}

Knowledge Graphs can be represented as multi-relational directed \textit{property graphs} \citep{Angles2018ThePG}, where nodes represent entities (for a general definition of \textit{entity}), and edges are predicates or relations. Any KG can thus be rendered as a list of triples ($subject$, $relation$, $object$)\footnote{several standards and formats exist for representing triples and optionally including additional metadata, including RDF, Turtle, N-triples, JSON-LD, and others.}, also called \textit{statements}\footnote{\url{https://www.wikidata.org/wiki/Help:Statements}}. 

An early line of work on knowledge graph construction focused on the TAC 2010 \textbf{Knowledge Base Population (KBP)} shared task \citep{ji2010overview}, which introduced a popular evaluation setting that separates knowledge base population into Entity Linking and Slot Filling subtasks. Early methods to address these tasks used pattern learning, distant supervision and hand-coded rules \cite{ji-grishman-2011-knowledge}. 


\textbf{Knowledge Graph Completion (KGC)} is a KG construction task that has gained popularity recently. It involves predicting missing links in incomplete knowledge graphs \cite{sun-etal-2020-evaluation}. The sub-tasks include \textbf{triple classification}, where models assess the validity of (head, relation, tail) triples; \textbf{link prediction}, which proposes subjects or objects for incomplete triples; and \textbf{relation prediction} \cite{shen2022comprehensive}, identifying relationships between subject and object pairs. Models for these tasks are frequently benchmarked against subsets of well-established knowledge bases such as WordNet \cite{fellbaum2010wordnet}, Freebase \cite{bollacker2008freebase}, and domain-specific KGs like UMLS \cite{bodenreider2004unified}. 

Evaluation methodologies for KG completion primarily utilize ranking-based metrics. These include Mean Rank (MR), Mean Reciprocal Rank (MRR), and Hits@K, which gauge a model's ability to prioritize correct triples over incorrect ones, offering a quantifiable measure of performance \cite{shen2022comprehensive}. 

Outside these tightly defined tracks, various approaches have been proposed to construct or populate knowledge graphs. For example, NELL (Never-Ending Language Learner) \cite{NELL-aaai15} is a self-supervised system that was designed to interact with the internet over years to populate a growing knowledge base of topical categories and factual statements.

\subsection{LLMs and Knowledge Graphs}

Studies have shown that pretrained language models (PLMs) possess factual and relational knowledge which makes them effective at downstream knowledge-intensive tasks such as open question-answering, fact verification, and information extraction \cite{shin2020autoprompt, petroni2019language}. KG-BERT \cite{yao2019kgbert} uses PLMs for KG completion by fine-tuning BERT on all KG completion sub tasks, treating the problem as a sequence classification task.

Pretrain-KG \cite{zhang-etal-2020-pretrain} introduces a framework that enriches knowledge graph embedding (KGE) models with PLM knowledge during training, which proves to be particularly useful for low-resource scenarios of link prediction and triple classification.

\paragraph{Knowledge Graph Construction Using Generative AI}

\begin{figure}[ht]
    \centering
    \resizebox{0.9\textwidth}{!}{%
        \begin{tikzpicture}[
    font=\ttfamily,
    square_node/.style args={#1#2}{
        rectangle, rounded corners, draw=none, fill=#1, 
        text width=#2, inner sep=2mm, align=left, font=\scriptsize,
    },
    square_node_header/.style args={#1#2#3#4}{
        rectangle, rounded corners, draw=none, fill=#1, 
        text width=#2, inner sep=1mm, align=left, font=\scriptsize,
        append after command={
            \pgfextra{
                \node [rectangle, draw, fill=white, text width=#3, align=center, font=\scriptsize, above=-0.5mm of \tikzlastnode.north] {#4};
            }
        },
    },    
]
    \tikzstyle{arrow} = [very thick,->,>=stealth, draw=black!50]
    

    \node[square_node={verylightgray}{4cm}] (input) {\tiny Tesla, Inc. announced on March 15, 2024, that its new electric vehicle model, the Tesla Cybertruck, will be launched in the United States in July 2024. Elon Musk, CEO of Tesla, stated that the Cybertruck represents a significant advancement in electric vehicle technology.};
    \node[square_node={customLightBlue}{4cm}, above=0.05cm of input] (input_prompt) {\tiny Extract all named entities and any relevant properties about them in the following text:};
    \begin{scope}[on background layer]
        \node (input_prompt_box) [square_node_header={customLightBlue}{2mm}{1.7cm}{User prompt}, fit=(input) (input_prompt)] {};
    \end{scope};

    \node (model) [square_node={customRed}{0.9cm}, right=1cm of input_prompt_box] {OpenIE System};    

    \node (output) [square_node_header={customLightBlue}{4cm}{3.3cm}{Open schema information}, right=1cm of model] {\tiny  
    ``entities": [\\
      \ \  \{\\
      \ \   \ \ ``Entity": "Tesla, Inc.",\\
      \ \  \ \``Type": ``Company",\\
      \ \   \ \ ``Action": ``Announced",\\
      \ \   \ \ ``Date": ``March 15, 2024"\\
      \ \  \},\\
      \ \  \{\\
      \ \   \ \ ``Entity": ``Tesla Cybertruck",\\
      \ \   \ \ ``Type": ``Electric Vehicle",\\
       \ \  \ \ ``Action": ``Launch",\\
       \ \  \ \ ``Location": ``United States",\\
      \ \   \ \ ``Date": ``July 2024"\\
      \ \  \},\\
      \ \  \{\\
      \ \   \ \ ``Entity": ``Elon Musk",\\
     \ \    \ \ ``Type": ``Person",\\
     \ \    \ \ ``Title": ``CEO",\\
     \ \    \ \ ``Affiliation": ``Tesla, Inc."\\
     \ \   \}\\
      ]\\
    };    

    \draw [arrow] (input_prompt_box) -- (model);
    \draw [arrow] (model) -- (output);

\end{tikzpicture}  
    }
    \caption{An example of Open Information Extraction. Note that in OpenIE, the output schema is not fixed.}
    \label{fig:example-openie}
\end{figure}
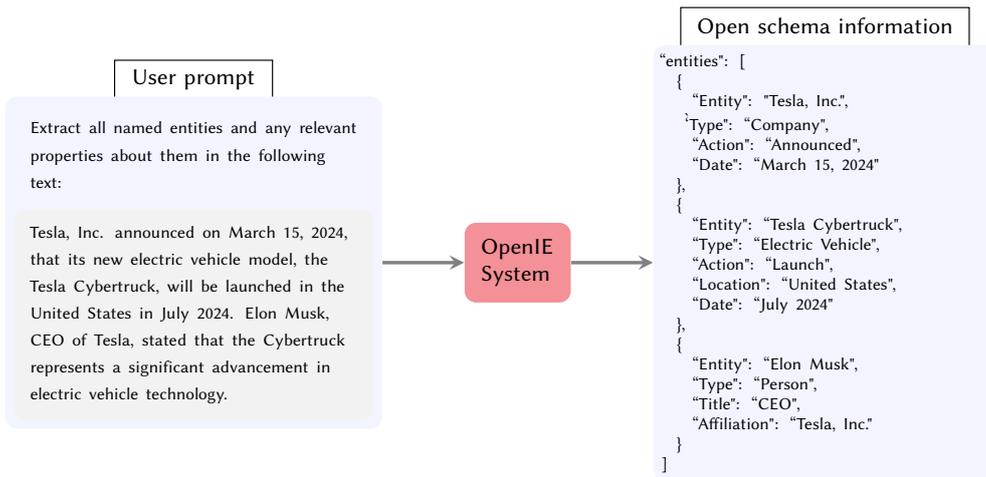

With the proliferation of general-purpose LLMs \cite{zhao2023survey}, open information extraction (OpenIE) has become one of the most popular industry applications of generative AI \cite{xu2023large}. OpenIE is closely related to knowledge graph construction, and so LLMs have naturally been applied to KG completion tasks such as link prediction and triple classification, proving to be successful in both fine-tuned \cite{zhang2023making} and zero-shot settings\citep{jiang2023structgpt,yao2024exploring}. The dominant paradigm is to include the desired schema of the output in the user prompt along with the input itself (refer to Figure \ref{fig:example-openie}). 



Khorashadizadeh et al. demonstrate the capabilities of GPT 3.5 in the task of KG construction using an in-context learning approach \cite{khorashadizadeh2023exploring}. Emphasis is placed on the importance of good prompt design under this setting. LLM2KB \cite{nayak2023llm2kb} fine-tunes open-source LLMs to predict tail entities given a head entity and relation, incorporating context retrieval from Wikipedia to enhance the relevance and accuracy of the predicted entities. \citet{zhu2024llms} investigate GPT-4's \cite{openai2024gpt4} capabilities for different steps of knowledge graph construction. They show that while GPT-4 exhibits modest performance on few-shot information extraction tasks, it excels as an inference assistant due to it's strong reasoning capabilities. Their experiments also show that GPT-4 generalizes well to new knowledge by creating a virtual knowledge extraction task.



Complementing these advancements, resources such as the Text2KG Benchmark \cite{mihindukulasooriya2023text2kgbench} offer valuable tools for researchers to develop and test LLM-backed KG completion models. This benchmark, specifically designed for evaluating knowledge graph generation from unstructured text using guideline ontologies, marks a significant step towards standardizing and accelerating research in this field.

 A comprehensive survey on the unification of LLMs and KGs \cite{Pan_2024} highlights the emergence of \textit{KG-enhanced LLMs}, \textit{LLM-augmented KGs}, and \textit{Synergized LLMs and KGs}. Validation and evaluation of KGs with LLMs has been less explored, but is also a promising and important avenue for research.

\subsection{Structuring and Validating Language Model Output}

Constraining language models to produce outputs that conform to specific schemas is challenging but essential for applications like natural language to SQL (NL2SQL) \cite{10.14778/3401960.3401970nl2sql, Guo2019Content}. Recent developments include tools like Guidance\footnote{\url{https://github.com/guidance-ai/guidance}}, Outlines\footnote{\url{https://github.com/outlines-dev/outlines}}, JSONFormer\footnote{\url{https://github.com/1rgs/jsonformer}}, and Guardrails\footnote{\url{https://github.com/guardrails-ai/guardrails}}, which facilitate constrained decoding of structured outputs from large language models (LLMs). Additionally, \textit{semantic validation} techniques like those enabled by the Instructor library use Pydantic classes to ensure outputs meet both structural and semantic accuracy. This advancement is crucial for tasks such as knowledge graph (KG) completion, where precision in data parsing significantly enhances model utility \cite{yao2024exploring}.

\subsection{Knowledge-Grounded LLMs}
\label{subsec:rag}

The tendency of LLMs to hallucinate poses a significant challenge in their application to downstream tasks \cite{Ji_2023hallucinate}. Retrieval-Augmented Generation (RAG) mitigates this by grounding LLM responses in verified information, significantly enhancing accuracy and reliability \cite{gao2024retrievalaugmented, semnani2023wikichat}. RAG integrates a retrieval component that leverages external knowledge during the generation process, improving performance across various natural language processing tasks \cite{lewis2021retrievalaugmented}. Additionally, role-playing approaches using LLMs have been developed to create detailed, organized content similar to Wikipedia articles, drawing on trusted sources for factual grounding \cite{shao2024assisting}.




\subsection{Knowledge Graph Evaluation}

Evaluating automatically constructed knowledge graphs is challenging. Huaman et al. present a comprehensive evaluation of state-of-the-art validation frameworks, tools, and methods for KGs \cite{huaman2020knowledge}. They highlight the challenges in validating KG assertions against real-world facts and the need for scalable, efficient, and effective semi-automatic validation approaches. \citet{gao2019efficient} have highlighted the trade-offs between human annotation cost and meaningful estimates of accuracy. As discussed above, a common flaw reported in existing KG evaluation frameworks is use of a closed-world assumption. Specifically, this means treating unknown predicted triples as false \cite{Mayfield2012EvaluatingTQ}. \citet{sun-etal-2020-evaluation} find that several recent KG completion techniques have reported significantly higher performance compared to earlier SoTA methods, in some cases due to the inappropriate evaluation protocols used. \citet{cao2021missing} suggest that triple classification evaluation under the closed-world assumption leads to trivial results. Additionally, Cao et al. note that current models lack the capacity to distinguish \textit{false} triples from \textit{unknown} triples. \citet{yang2022rethinking} confirm the existing gap between closed and open world settings in the performance of KG completion models.

\section{Approach}
\label{sec:approach}

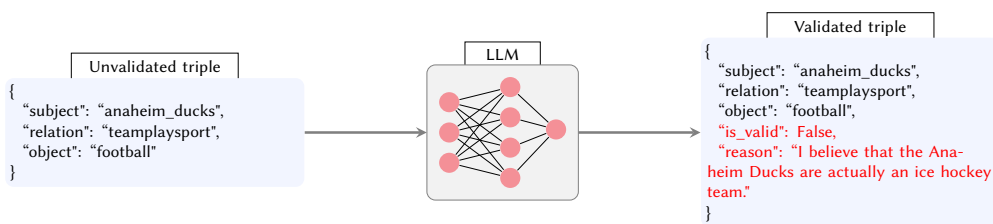
\begin{figure}[ht!]
    \centering
    \resizebox{0.9\textwidth}{!}{%
        \begin{tikzpicture}[
    font=\ttfamily,
    bigbox/.style args={#1#2#3#4}{
        rounded corners, fill=#1, draw=gray, align=center, inner sep=#2, line width=0.1pt,
        append after command={
            \pgfextra{
                \node [rectangle, draw, fill=white, text width=#3, align=center, font=\scriptsize, above=-0.5mm of \tikzlastnode.north] {#4};
            }
        }
    },
    square_node/.style 2 args={
        rectangle, rounded corners, draw=none, fill=#1, 
        text width=#2, minimum height=#2, inner sep=1mm, align=center, font=\scriptsize,
    },
    square_node_header/.style 2 args={
        rectangle, rounded corners, draw=none, fill=#1, 
        text width=4.7cm, inner sep=1mm, align=left, font=\scriptsize,
        append after command={
            \pgfextra{
                \node [rectangle, draw, fill=white, text width=2.5cm, align=center, font=\scriptsize, above=-1mm of \tikzlastnode.north] {#2};
            }
        }
    },
    neuron/.style={circle, draw=none, fill=customRed, minimum size=0.25cm},
    pics/llm/.style={
        code={
            \begin{scope}[local bounding box=llm]
              \foreach \i in {1,...,3}
                \node[neuron] (-input-\i) at (0,\i*0.5-1) {};
              \foreach \i in {1,...,4}
                \node[neuron] (-hidden-\i) at (1,\i*0.5-1.25) {};
              \node[neuron] (-output) at (1.75,0.05) {};
              \foreach \i in {1,...,3}
                \foreach \j in {1,...,4}
                  \draw (-input-\i) -- (-hidden-\j);
              \foreach \i in {1,...,4}
                \draw (-hidden-\i) -- (-output);
            \end{scope}
        }
    }
]
    \tikzstyle{arrow} = [very thick,->,>=stealth, draw=black!50]
    \tikzstyle{arrow_dashed} = [very thick, dashed, ->, >=stealth, draw=black!50]
    

    \pic {llm};

    \begin{scope}[on background layer]
        \node (external_llm_box) [bigbox={verylightgray}{2mm}{1cm}{LLM}, fit=(llm)] {};
    \end{scope};

    \node (unvalidated_triple1) [square_node_header={customLightBlue}{Unvalidated triple}, left=2cm of external_llm_box] {
        \{\\
        \ \ ``subject": ``anaheim\_ducks",\\
        \ \ ``relation": ``teamplaysport",\\
        \ \ ``object": ``football"\\
        \}\\
    };
    
    \node (validated_triple1) [square_node_header={customLightBlue}{Validated triple}, right=2cm of external_llm_box] {
        \{\\
        \ \ ``subject": ``anaheim\_ducks",\\
        \ \ ``relation": ``teamplaysport",\\
        \ \ ``object": ``football",\\
        \ \ \textcolor{red}{``is\_valid": False,}\\
        \ \ \textcolor{red}{``reason": ``I believe that the Anaheim Ducks are actually an ice hockey team."}\\
        \}    
    };

    \draw [arrow] (unvalidated_triple1) -- (external_llm_box);
    \draw [arrow] (external_llm_box) -- (validated_triple1);

\end{tikzpicture}  
    }
    \caption{Validating KGs with LLM Knowledge}
    \label{fig:3-1-a}
\end{figure}

We assume the existence of a triple-extractor model, which produces a stream of candidate \textit{statements} from unstructured data feeds. The triple-extractor model could be implemented by a KG completion model, one or more LLMs with well-designed prompts, or by a more traditional information extraction pipeline consisting of several distinct models that perform parsing, named entity recognition, relationship classification, and other relevant sub-tasks. For each predicted triple from the stream, we wish to validate whether it is correct in the presence of context. Once a statement has been validated, it can be written into a knowledge graph or another data store, and statements that do not pass validation can be flagged for further review. A high-level overview of the validation stage is illustrated in Figure \ref{fig:3-1-a}.

\noindent In this work we use existing standard KGC datasets for our experiments, so in practice the candidate triples in this work are produced by streaming through existing datasets (see Section \ref{sec:experiments}).

Possible sources of context for validation include:

\begin{itemize}
    \item Knowledge accrued in the LLM parameters during pretraining.
    \item User-provided context in the form of document collections or reference KGs represented in string format.
    \item Agents that can interact with the world to search and retrieve information in various ways.
\end{itemize}

Further detail on use of context in our validator implementations is provided in Section \ref{context-for-val}.

\paragraph{Basic Settings for Validation:} The first step is to obtain KG completion predictions in the format of a list of ($h$, $r$, $t$) triples, each consisting of a head entity $h$, a relation $r$ and a tail entity $t$. All validators are instantiated in a zero-shot setting with an LLM backbone; this may be a model from OpenAI's model family, such as gpt-3.5-turbo-0125 \cite{gpt3, openai2024gpt4}, or an open-source model from the Llama family \cite{touvron2023llama}. Additionally, validators have access to various tools which allow them to query external knowledge sources.

\paragraph{Validation via Pydantic Models}
Pydantic is a data validation and settings management library which leverages Python type annotations. It allows for the creation of data models, where each model defines various fields with types and validation requirements. By using Python's type hints, Pydantic ensures that the incoming data conforms to the defined model structure, performing automatic validation at runtime.

KG triples are passed to the validator via the Instructor library, which uses a patched version of popular LLM API clients. This patch enables the request of structured outputs in the form of Pydantic classes. It is within these Pydantic classes that we specify the structural and semantic guidelines that the LLM must follow during validation. An example of this form of prompting is shown in Figure \ref{fig:ValidatedTriple}. Specifically, we request that, for every triple $(h, r, t)$, the model must provide values for a number of fields:

\begin{enumerate}
    \item \texttt{triple is valid}: A boolean indicating whether the proposed triple is generally valid, judged against any given context. The model can reply with  \texttt{True}, \texttt{False}, or \texttt{"Not enough information to say"}.
    \item \texttt{reason}: An open-form string describing why the triple is or is not valid.
\end{enumerate}


\subsection{Validation Contexts}\label{context-for-val}

This section discusses the contextual information that is available to different validator instantiations. We use \textit{context} to mean all information that is available to a validator, including the information stored in trained model parameters.

\subsubsection{Validating with LLM Knowledge}
\label{without context}
This is the most straightforward method of triple validation. Given a triple ($h$, $r$, $t$), the objective is to classify the triple using the LLM's inherent knowledge about the world, learned during the pretraining stage, and stored in the model parameters. The process is illustrated in Figure \ref{fig:3-1-a} and an example can be found in appendix Figure \ref{fig:no-context}. This is a powerful and simple way to verify triples with no additional data.

\subsubsection{Validation using Textual Context(s)}
\label{text-context}
\begin{figure}[htbp]
  \centering
  \resizebox{\textwidth}{!}{%
    \includegraphics{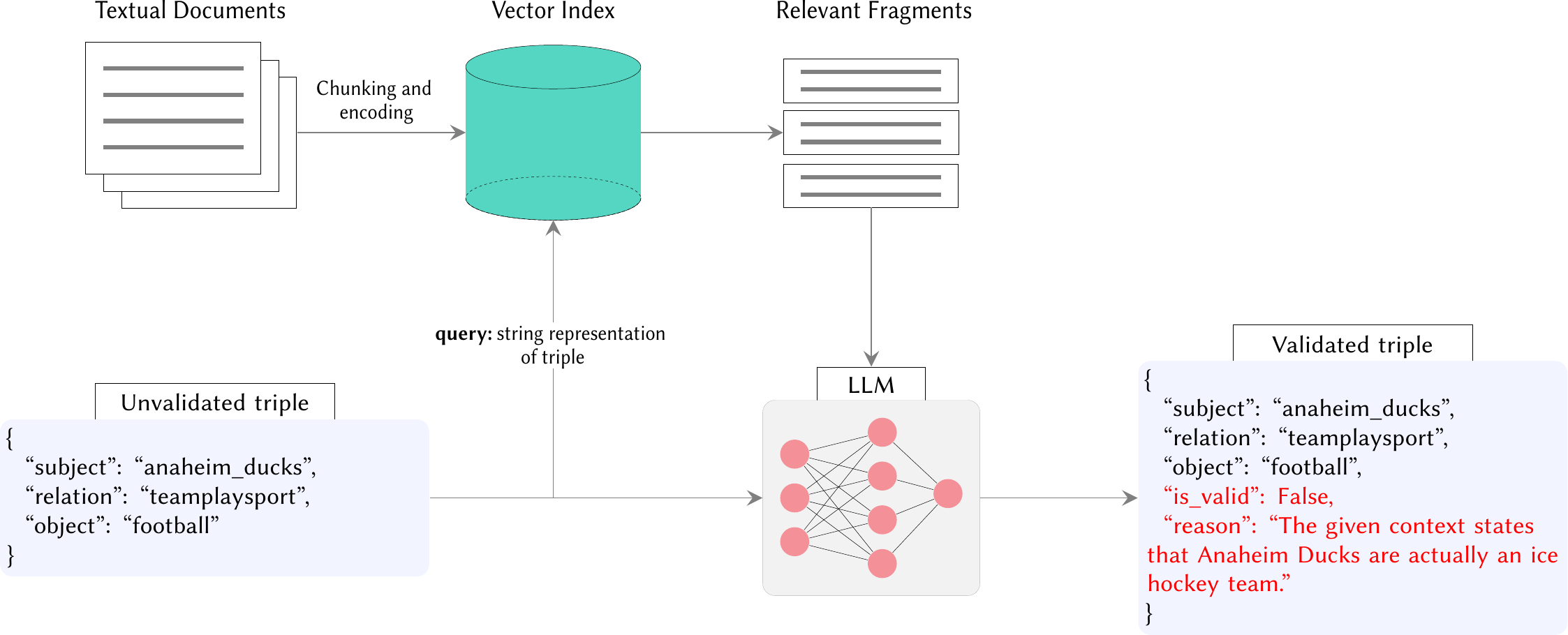}
  }
  \caption{Validating KGs given Textual Context}
  \label{fig:3-1-b}
\end{figure}

Inspired by the success of Retrieval-Augmented Generation (RAG) in knowledge-intensive tasks such as question-answering \cite{gao2024retrievalaugmented}, we implement tooling to retrieve relevant information from a reference text corpus (see Section \ref{subsec:rag}). In this instance, the model is prompted with textual context alongside the candidate triple, as shown in Figure \ref{fig:3-1-b}. This approach is particularly useful for a number of scenarios:

\begin{itemize}
    \item When we wish to verify a set of triples about the same entity or group of entities and we have a collection of trustworthy sources within which we assume there will be evidence for or against the predicted triple, for example a given entity's Wikipedia page.
    \item  When building KGs using private or domain-specific data feeds.
\end{itemize} 

\noindent This provided corpus can be of arbitrary length and can contain a collection of documents. The corpus will be recursively chunked and encoded by an embedding model from either the sentence transformers library \cite{reimers-2019-sentence-bert}  or OpenAI's family of embedding models \cite{neelakantan2022text}, and a searchable index is created. A string representation for each triple is then constructed, and this is used to query the corpus index, which retrieves the most semantically similar chunks of text, according to cosine similarity. This forms the context against which the LLM will validate the given triple.

\subsubsection{Validation using a Reference KG}
\label{wikidata-context}

We also consider validating proposed KG triples by cross-referencing against established, reliable KGs. Wikidata, with its expansive and well-structured repository of knowledge, serves as an ideal reference point for such validations, and will serve as the reference KG in our experiments. However, we note that any KG can be used as a reference by following the method outlined in \ref{ref-kg-method}. 

The Wikidata knowledge graph is built from two top-level types: \textit{Entities} and \textit{Properties}:

\paragraph{Entities:} Entities represent all items in the database. An item is a real-world object, concept, or event, such as ``Earth" (Q2), ``love" (Q316), or ``World War II" (Q362). Items can be linked to each other to form complex statements via properties. In the context of KG completion, a statement can be thought of as a triple. Each entity is identified by a unique identifier, which is a Q-prefix followed by a sequence of numbers, e.g., Q42 for Douglas Adams.

\paragraph{Properties:} Properties in Wikidata define the characteristics or attributes of items and establish relationships between them. They are the predicates in statements, linking subjects (items) to their object (value or another item). For example, in the statement ``Douglas Adams (Q42) - profession (P106) - writer (Q36180)",``profession" is the property that describes the relationship between ``Douglas Adams" and ``writer".


\paragraph{Reference KG Implementation}\label{ref-kg-method} Our approach to integrating Wikidata as a source of contextual information is simple. Given triple $t$, an agent module searches Wikidata using the string of the subject as a query. The top Wikidata entity from the search API is returned -- if no results are found for the query, a warning is thrown, and the validator will default to using its inherent knowledge. The Wikidata item is parsed to remove a list of trivial properties. Among Wikidata's 11,000 Properties, over 7,000 of these are identifiers to external databases such as IMDb and Reddit \footnote{\url{https://wikiedu.org/blog/2022/03/30/property-exploration-how-do-i-learn-more-about-properties-on-wikidata/}}. In this work, we are not interested in verifying such information, and so we discard these properties.

A string representation of the Wikidata page is now passed through the same RAG pipeline as described in Section \ref{text-context}, from which relevant sections are retrieved and passed to the validator as context alongside each predicted triple $t$. This implementation is illustrated in appendix Figure \ref{fig:3-1-c}.

\subsection{Validation using Web Search}
In some cases, the triples we wish to validate cannot be captured with a query to Wikidata, and we do not have a collection of textual information to provide the model with additional context. To overcome this, the validator is given access to collect information relevant to the triple via a web-searching agent. The triple is formatted as a string query. An agent then searches the web using the DuckDuckGo API\footnote{\url{https://github.com/deedy5/duckduckgo_search}}. The top results for the given query are parsed and stored as a collection of documents. The validation then follows the same pattern as Section \ref{text-context}, whereby relevant chunks of text are retrieved as context for triple validation. This method is illustrated in appendix Figure \ref{fig:3-2}.

\section{Experiments}
\label{sec:experiments}

We conduct a series of triple classification experiments to validate the effectiveness of an LLM-backed validator for KG Completion. Our experiments make use of a number of popular benchmark KG datasets: UMLS \cite{bodenreider2004unified}, WN18RR \cite{dettmers2018convolutional}, FB15K-237N, Wiki27k \cite{lv-etal-2022-pre}, and CoDeX-S \cite{safavi-koutra-2020-codex}. FB15k-237N is derived from Freebase, and was obtained by removing the relations containing mediator nodes in FB15K-237. Wiki27K was created from Wikidata and manually annotated with real negative triples. UMLS is a medical ontology describing relations between medical concepts. WN18RR is a dataset about English morphology derived from WordNet. We investigate the performance of \texttt{gpt-3.5-turbo-0125} and \texttt{gpt-4-0125-preview} and present our results in Tables \ref{tab:fb15k-wiki27k} , \ref{tab:wn18rr-umls} and \ref{tab:codex}. Setup details and results for open-source LLM experiments can be found in Section \ref{oss-experiments} and Table \ref{tab:llama-2} in the appendix.

\begin{table}[htbp!]
\caption{Experiment results for FB15K-237N-150 and Wiki27K-150 datasets. Accuracy (Acc), precision (P), recall (R), and F1-score (F1) results for each method are reported. The best metrics for each dataset are marked in bold.}
\label{tab:fb15k-wiki27k}
\centering
\begin{tabular}{@{}lllllllll@{}}
\toprule
\multirow{2}{*}{Model} & \multicolumn{4}{c}{FB15K-237N-150} & \multicolumn{4}{c}{Wiki27K-150} \\ \cmidrule(l){2-9} 
                       & P      & R      & F1     & Acc    & P     & R      & F1     & Acc    \\
GPT 3.5 WorldKnowledge & 0.58   & \textbf{0.97}   & 0.73   & 0.63   & 0.63  & \textbf{1.0}    & 0.77   & 0.71   \\
GPT 3.5 Wikidata       & 0.75   & 0.77   & 0.76   & 0.76   & 0.74  & 0.73   & 0.74   & 0.74   \\
GPT 3.5 WikipediaWikidata & 0.85  & 0.69   & 0.76   & 0.79   & 0.84  & 0.86   & 0.85   & 0.85   \\
GPT 3.5 Web            & 0.76   & 0.85   & 0.81   & 0.79   & 0.76  & 0.91   & 0.82   & 0.81   \\
GPT 3.5 WikidataWeb    & 0.82   & 0.81   & \textbf{0.82}   & 0.82   & 0.78  & 0.87   & 0.82   & 0.81   \\
GPT 4 WorldKnowledge & 0.87& 0.72& 0.79& 0.81& 0.95& 0.76& 0.84& 0.86\\
GPT 4 Wikidata       & 0.89& 0.64& 0.74& 0.78& 0.97& 0.75& 0.84& 0.86\\
GPT 4 WikipediaWikidata & 0.90& 0.59& 0.71& 0.76& 0.97& 0.77& 0.86& 0.87\\
GPT 4 Web            & \textbf{0.92}& 0.72& 0.81& \textbf{0.83}& 0.95& 0.75& 0.84& 0.85\\
GPT 4 WikidataWeb    & \textbf{0.92}& 0.72& 0.81& \textbf{0.83}& \textbf{1.0}& 0.77& \textbf{0.87}& \textbf{0.89}\\
\bottomrule
\end{tabular}
\end{table}

\begin{table}[htbp!]
\caption{Experiment results for WN18RR-150 and UMLS-150 datasets. Accuracy (Acc), precision (P), recall (R), and F1-score (F1) results for each method are reported. The best metrics for each dataset are marked in bold.}
\label{tab:wn18rr-umls}
\centering
\begin{tabular}{@{}lllllllll@{}}
\toprule
\multirow{2}{*}{Model} & \multicolumn{4}{c}{WN18RR-150} & \multicolumn{4}{c}{UMLS-150} \\ \cmidrule(l){2-9} 
                       & P      & R      & F1     & Acc     & P     & R      & F1     & Acc    \\
GPT 3.5 WorldKnowledge & 0.54& 0.97& 0.70& 0.58 & 0.5& \textbf{0.97}& 0.66& 0.5\\
GPT 3.5 Wikidata       & 0.53& \textbf{0.99}& 0.69& 0.56 & 0.51& 0.87& 0.64& 0.52\\
GPT 3.5 WikipediaWikidata & 0.54& \textbf{0.99}& 0.69& 0.57 & 0.53& 0.88& 0.66& 0.55\\
GPT 3.5 Web            & 0.67& 0.97& 0.79& 0.74 & 0.52& 0.93& \textbf{0.67}& 0.53\\
GPT 3.5 WikidataWeb    & 0.69& 0.95& 0.80& 0.76 & 0.5& 0.88& 0.64& 0.5\\
GPT 4 WorldKnowledge & 0.99& 0.92& \textbf{0.95}& \textbf{0.95}& 0.57& 0.77& 0.66& 0.59\\
GPT 4 Wikidata       & 0.99& 0.91& 0.94& \textbf{0.95}& \textbf{0.63}& 0.69& 0.66& \textbf{0.64}\\
GPT 4 WikipediaWikidata & 0.99& 0.91& 0.94& \textbf{0.95}& 0.62& 0.67& 0.64& 0.63\\
GPT 4 Web            & \textbf{1.0}& 0.89& 0.94& \textbf{0.95}& 0.61& 0.65& 0.63& 0.62\\
GPT 4 WikidataWeb    & \textbf{1.0}& 0.88& 0.94& 0.94& 0.56& 0.64& 0.60& 0.57\\
\bottomrule
\end{tabular}
\end{table}

\begin{table}[htbp!]
\caption{Experiment results for CoDeX-150 dataset. Accuracy (Acc), precision (P), recall (R), and F1-score (F1) results for each method are reported. The best metrics are marked in bold.}
\label{tab:codex}
\centering
\begin{tabular}{@{}lllll}
\toprule
\multirow{2}{*}{Model}  &   \multicolumn{4}{c}{CoDeX-S-150}\\ \cmidrule(l){2-5} 
                       & P      & R      & F1     &Acc    \\
GPT 3.5 WorldKnowledge & 0.52& \textbf{0.97}& 0.68&0.54\\
GPT 3.5 Wikidata       & 0.86& 0.88& 0.87&0.87\\
GPT 3.5 WikipediaWikidata & 0.81& 0.87& 0.84&0.83\\
GPT 3.5 Web            & 0.74& 0.84& 0.79&0.77\\
GPT 3.5 WikidataWeb    & 0.87& \textbf{0.97}& \textbf{0.92}&\textbf{0.91}\\
GPT 4 WorldKnowledge & 0.87& 0.81& 0.84&0.85\\
GPT 4 Wikidata       & 0.93& 0.87& 0.90&0.9\\
GPT 4 WikipediaWikidata & \textbf{0.94}& 0.83& 0.88&0.89\\
GPT 4 Web            & 0.85& 0.84& 0.85&0.85\\
GPT 4 WikidataWeb    & 0.93& 0.85& 0.89&0.89\\
\bottomrule
\end{tabular}

\end{table}
\subsection{Experiment Settings}

\paragraph{Prompt as a Hyperparameter: } 
We emphasize the notion of a prompt as a model hyperparameter, and manually tuning it to fit a subset of data is a form of over-fitting or evaluation set leakage. In this work, we thus formulate a generic model prompt, and apply this prompt to all benchmark datasets without further changes. We include the prompt in the appendix (see Figure \ref{fig:prompt-used}).

Through the following experiments we attempt to answer the question: 
Given context, can our model judge whether an unseen triple ($h$, $r$, $t$) is correct? 

We are primarily interested in observing the change in evaluation performance of an LLM when it has access to context under the following settings:
\begin{itemize}
    \item \textbf{LLM Inherent Knowledge}: Evaluates the model's native understanding without external data sources.
    \item \textbf{Wikidata}: Uses structured data from Wikidata as the reference KG context.
    \item \textbf{Web}: Incorporates information retrieved directly from the internet.
    \item \textbf{WikidataWeb}: Combines data from both Wikidata and web sources.
    \item \textbf{WikipediaWikidata}: Utilizes a mix of Wikipedia and Wikidata to provide a comprehensive context.
\end{itemize}

\paragraph{API Cost and Rate-Limiting Constraints}
Due to OpenAI API constraints, we run experiments using a subset of 150 examples from each dataset. This is indicated by the \texttt{-150} suffix to each dataset name.










\section{Discussion}
\label{sec:discussion}

\subsection{Analysis}


Our analysis reveals notable variations in performance across datasets, as evidenced by the results obtained using different validators powered by GPT-3.5 and GPT-4 language models. Specifically, the GPT-3.5 \texttt{World Knowledge} validator shows limited effectiveness on the FB15K-237N-150, Wiki27k-150, and CoDeX-S-150 datasets (as detailed in Tables \ref{tab:fb15k-wiki27k} and \ref{tab:codex}). However, the introduction of contextual information from Wikidata and web searches gives a strong performance boost, with the performance on the CoDeX-S-150 dataset in particular improving accuracy from 0.54 to 0.91 when using the \texttt{WikidataWeb} validator.

GPT-4 configurations exhibit strong performance across the board, particularly excelling in the FB15K-237N-150 and Wiki27k-150 datasets, where GPT-4 achieves the highest accuracy of 0.83 and 0.89 respectively. However, both GPT-3.5 and GPT-4 models demonstrate less satisfactory results on the UMLS-150 dataset, as indicated in Table \ref{tab:wn18rr-umls}.

It is noteworthy that the incorporation of context from external knowledge sources, especially web searches and Wikidata, proves beneficial for both models. Despite this, the open-source Llama2 model performs poorly on this task, as shown in Table \ref{tab:llama-2} and inference examples \ref{fig:llama-2-example-1} and \ref{fig:llama-2-example-2}. We hypothesize that future open-source LLMs may perform much better than the those currently available.

GPT-4 validators display effectiveness on the WN18RR-150 dataset, both with and without supplemental context. This robust performance is hypothesized to stem from the model's superior grasp of English morphology and nuanced language comprehension, aligning with the linguistic focus of the WN18RR dataset.

\subsection{Key Findings}

\paragraph{Inherent Knowledge Insufficiency:} In the case of GPT-3.5 and Llama2 70B, reliance solely on the inherent knowledge of LLM validators often leads to unquestioned acceptance of predicted triples. This indicates a limitation in the models' ability to challenge the veracity of the information, underscoring the need for external validation mechanisms. This corroborates with findings from prior studies which find that LLMs struggle to memorize knowledge in the long tail \cite{mallen2023trust}.

\paragraph{Challenge in Verifying Ambiguous Triples:} Our evaluation of each dataset reveals that additional information is neccesary to verify many triples. For example, a positive triple in the UMLS dataset reads \texttt{("age\_group", "performs", "social\_behavior")}. Ambiguous triples in the UMLS and WN18RR datasets require understanding of specific ontologies, rendering web or Wikidata searches ineffective for retrieving relevant context. This complexity is contrasted by datasets like FB15K-237N and Wiki27k, which involve concrete entities or facts (e.g., people, locations) more amenable to validation through widely available external sources. For example, a positive example in FB15K-237N reads \texttt{{"Tim Robbins", "The gender of [X] is [Y] .", "male"}},

\paragraph{The Importance of Relevant Context:} Performance is weaker on datasets requiring domain-specific knowledge, such as UMLS, where no model tested achieved satisfying results. This is attributed to the challenge of sourcing pertinent context for validation, as exemplified by the clinical domain triple from UMLS: \texttt{("research\_device", "causes", "anatomical\_abnormality")}. This highlights the critical role of context in enabling accurate validation, emphasizing the need for targeted search strategies to augment the model's knowledge base.

\paragraph{Limitations in Zero-Shot Triple Classification by Current Open-Source LLMs:} Table \ref{tab:llama-2} shows the performance of a version of the LLama-2-70B-chat model \cite{touvron2023llama} on the triple verification task. Upon manual inspection, the model nearly always returns a \texttt{True} prediction for all triples, irrespective of the provided context, resulting in a recall of 1.0 and precision of about 0.5 across all settings. This tendency suggests that while the model can superficially engage with the context—evident from relevant factoids appearing in the \texttt{reason} field—it often resorts to fabricating agreeable responses rather than accurately assessing the triple's validity. Figures \ref{fig:llama-2-example-1} and \ref{fig:llama-2-example-2} illustrate this behaviour. It is worth noting that our experiments were conducted using a single open-source model; however, alternative models could potentially deliver superior performance. We propose this as an avenue for future research. 

\paragraph{Adoption of Other Open-Source LLMs: } At present, we find that only OpenAI and Llama models are usable with the Instructor framework. More recent models, such as Mixtral \cite{jiang2024mixtral} and Gemma \cite{gemmateam2024gemma}, are beginning to receive support under this library, but issues with constraining model output has delayed implementation. We are particularly interested in observing how other open-source models perform at this task in the future.





\subsection{Ethical and Social Risks}

Building on the framework by \citet{weidinger_ethical_2021}, we highlight key ethical and social risks associated with using LLMs for KG validation. LLMs, trained on large-scale internet datasets, may perpetuate biases \cite{bender_dangers_2021}, discriminating against marginalized groups and potentially reinforcing stereotypes within KGs. Additionally, the alignment of LLM outputs with human preferences can introduce biases favoring certain languages and perspectives \cite{ryan_unintended_2024}. Privacy concerns also arise from LLMs potentially leaking sensitive information \cite{carlini_extracting_2021}. Furthermore, the risk of spreading misinformation through inaccurate validation poses serious challenges, especially in sensitive domains like medicine or law. Lastly, the environmental impact of training and deploying LLMs, including significant carbon emissions and water usage, underscores the need for sustainable practices in LLM-driven KG validation \cite{mytton_data_2021, patterson_carbon_2021}.

\section{Conclusions}
\label{sec:conclusions}

We have introduced a flexible framework for utilizing large language models for the validation of triples within knowledge graphs, capitalizing on both the inherent knowledge embedded within these models, and upon supplementary context drawn from external sources. As demonstrated in our experiments (Section \ref{sec:experiments}), the approach significantly enhances the accuracy of zero-shot triple classification across several benchmark KG completion datasets, provided that the appropriate context can be retrieved from external sources. 

\paragraph{Use Cases:} From experimentation, LLMs have demonstrated the potential to be effective validators for KG completion methods. They also open up the possibility of updating existing KG datasets with new knowledge from external sources, ensuring their relevance as gold-standard benchmarks. A practical application of this is the development of automated systems, such as bots, designed to enrich platforms like Wikidata with real-world data. These bot contributions could be systematically verified by SoTA LLMs to ensure accuracy and relevance.

As of January 2024, Wikidata encompassed nearly 110 million pages, a figure increasing at an accelerating rate. The decade between 2014 and 2023 saw an annual average of 9.57 million new pages and 191.5 million edits, and cumulative annual growth rates of 12.83\% and 12.16\% respectively \cite{wikimedia_wikidata_stats_2024}. The volume and pace of such expansion highlights the challenge of relying on manual verification methods. Leveraging LLMs to flag incorrect or unsupported edits made by users or bots could be an excellent aid to the Semantic Web community.

\paragraph{Future Research:} As the quality of general-purpose LLMs improves, this framework should become increasingly effective in validating KG completion models. Instructor has already begun work to support other open-source LLMs, which would enable even greater flexibility in validator configuration. 

Enriching models with domain-specific context and graph structural features could boost their performance across diverse datasets. Moreover, fine-tuning strategies tailored to LLMs may unlock even better performance when a model is fine-tuning specifically for the KG validation task. 

As discussed in Sections \ref{sec:introduction} and \ref{sec:background}, a growing body of work studies knowledge graph creation and augmentation using generative models. Knowledge graph creation is outside the scope of this paper, but we plan to explore this in future work. Given an information extraction model which produces KG triples from raw text, our verification pipeline could be connected to the entity and property stores of an existing KG, and automatically update the KG with high-accuracy information extracted from textual data feeds such as news. We note this is likely to be easier for some domains then others, and current SoTA LLMs will probably not be good verifiers for domain specific KGs.

\bibliography{references}

\newpage
\appendix

\section{Appendix}

\subsection{Prompt Templates}

\begin{figure}[ht]

\begin{minted}[frame=single,
               framesep=1mm,
               fontsize=\footnotesize,
               breaklines]{python}
@staticmethod
def validate_statement_with_no_context(entity_label, predicted_property_name, predicted_property_value):
    '''Validate a statement about an entity with no context

    a statement is a triple: entity_label --> predicted_property_name --> predicted_property_value
                            e.g Donald Trump --> wife --> Ivanka Trump
    
    '''
    resp: ValidatedTriple = client.chat.completions.create(
        response_model=ValidatedTriple,
        messages=[
            {
                "role": "user",
                "content": f"Using your vast knowledge of the world, " +
                        "evaluate the predicted Knowledge Graph triple for its accuracy by considering:\n" +
                        "1. Definitions, relevance, and any cultural or domain-specific nuances of key terms\n" + 
                        "2. Historical and factual validity, including any recent updates or debates around the information\n" + 
                        "3. The validity of synonyms or related terms of the prediction\n" + 
                        "Approach this with a mindset that allows for exploratory analysis and the recognition of uncertainty or multiple valid perspectives. " +
                        "Use this approach to recognize a range of correct answers when nuances and context allow for it." +
                        "If multiple relations are provided, the triple is valid if any of them are valid. " +
                        f"\nSubject Name: {entity_label}" + 
                        f"\nRelation: {predicted_property_name}" + 
                        f"\nObject Name: {predicted_property_value}"
            }
        ],
        max_retries=3,
        temperature=0,
        model=MODEL,
    )
    return resp
\end{minted}
\vspace{-10pt}
\captionof{figure}{The prompt used across all experiments. The LLM response is captured as a Pydantic model.}
\label{fig:prompt-used}

\end{figure}

\begin{figure}[ht]

\begin{minted}[frame=single,
               framesep=1mm,
               fontsize=\footnotesize,
               breaklines]{python}
class ValidatedTriple(BaseModel, extra='allow'):
    predicted_subject_name: str
    predicted_relation: Union[str, List[str]]
    predicted_object_name: str

    reason: str = Field(
        ..., description="The reason why the predicted subject-relation-object triple is or is not valid."
    )
    triple_is_valid: Literal[True, False, "Not enough information to say"] = Field(
      ...,
        description="Whether the predicted subject-relation-object triple is generally valid, following the previously-stated approach. " +
                    "If multiple relations are provided, the triple is valid if any of them is valid. " +
                    "Think through the context and the nuances of the terms before providing your answer. " +
                    "If the context does not provide enough information, try to use your common sense."
    )
\end{minted}
\vspace{-10pt}
\captionof{figure}{The Pydantic model which will encapsulate the LLM Validator response.}
\label{fig:ValidatedTriple}

\end{figure}

\newpage

\subsection{Validators with Context Illustrations}

\begin{figure}[htbp]
  \centering
  \resizebox{\textwidth}{!}{%
    \includegraphics{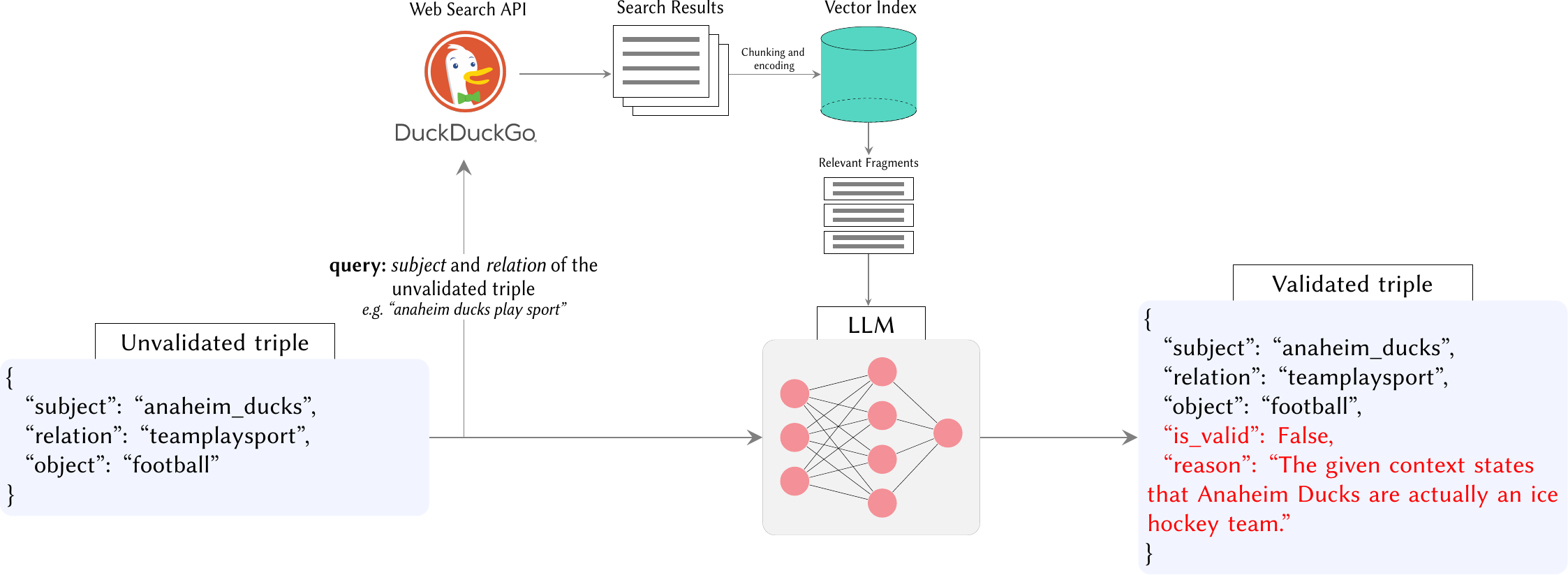}
  }
  \caption{Validating KGs using Web Search}
  \label{fig:3-2}
\end{figure}

\begin{figure}[htbp]
  \centering
  \resizebox{\textwidth}{!}{%
    \includegraphics{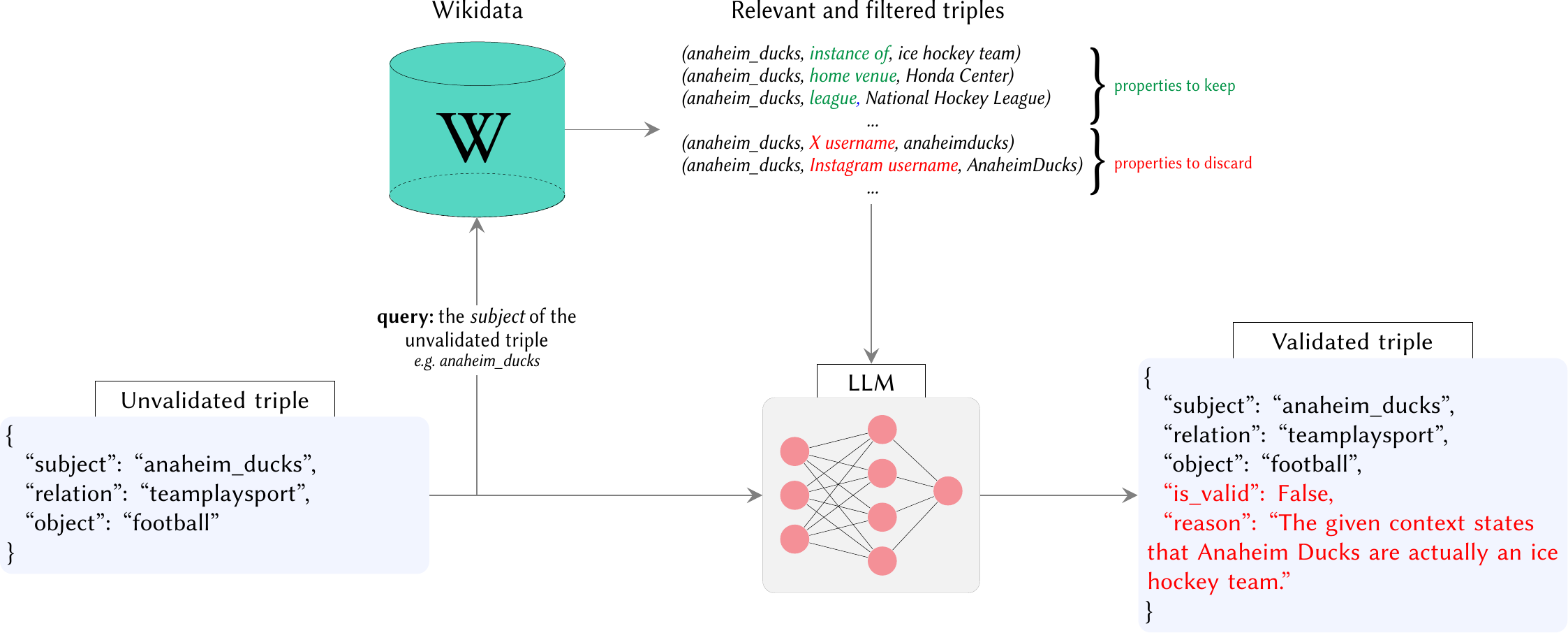}
  }
  \caption{Validating KGs given a Reference KG such as Wikidata}
  \label{fig:3-1-c}
\end{figure}

\FloatBarrier

\subsection{Open-source Experimental Setup}
\label{oss-experiments}

To evaluate the capabilities of open-source LLMs in our
framework, we employed a version of the LLama-2-70B-chat model \cite{touvron2023llama}. We selected a model applying Q5\_K\_M quantization to LLama2 70b-chat, chosen for its minimal quantization levels and low reported impact on quality, provided by \cite{the_bloke_model}. To implement this, we take advantage of Instructor's integration with llama-cpp-python\footnote{\url{https://github.com/abetlen/llama-cpp-python}}, which supports quantized models in GGUF \cite{ggerganov2023gguf} format available on Hugging Face Hub\footnote{\url{https://huggingface.co/docs/hub}}. Our experimental setup for open-source LLMs uses two NVIDIA A100 GPUs. 

\subsection{Open-source LLM Experimental Results}

\begin{table}[ht]
\caption{Experiment results using Llama-2-70B-chat model for FB15K-237-N-150, CoDeX-S-150 and Wiki27K-150 datasets. Accuracy (Acc), precision (P), recall (R), and F1-score (F1) results for each method are reported. Scores for each dataset are separated by slashes, listed as FB15K-237-N-150/CoDeX-S-150/Wiki27K-150.}
\label{tab:llama-2}
\centering
\begin{tabular}{lcccc}
\toprule
Model & P & R & F1 & Acc \\
\midrule
Llama-2 Web               & 0.52/0.51/0.46 & 1.0/1.0/1.0 & 0.68/0.67/0.63 & 0.54/0.52/0.49 \\
Llama-2 WorldKnowledge    & 0.54/0.51/0.54 & 1.0/1.0/1.0 & 0.70/0.66/0.70 & 0.58/0.50/0.58 \\
Llama-2 Wikidata          & 0.53/0.55/0.53 & 1.0/1.0/1.0 & 0.69/0.71/0.69 & 0.56/0.60/0.56 \\
Llama-2 WikidataWeb       & 0.50/0.50/0.51 & 1.0/1.0/1.0 & 0.66/0.66/0.67 & 0.50/0.50/0.51 \\
Llama-2 WikipediaWikidata & 0.51/0.50/0.51 & 1.0/1.0/1.0 & 0.67/0.66/0.67 & 0.51/0.50/0.51 \\
\bottomrule
\end{tabular}
\end{table}

\FloatBarrier
\subsection{Validator Inference Examples}

\begin{figure}
\begin{minted}[frame=single,
               framesep=2mm,
               baselinestretch=1.2,
               fontsize=\footnotesize,
               breaklines]{python}
>>> inp = [{"predicted_subject_name": "Heinrich Rudolf Hertz",
       "predicted_relation": [
        "occupation"
      ],
      "predicted_object_name": "theologian",}
    ]
>>> WorldKnowledgeKGValidator(**{'triples': inp})
>>> {
      "triple_is_valid": false,
      "reason": "Heinrich Rudolf Hertz was a German physicist who made significant contributions to the field of electromagnetism and is best known for his discovery of electromagnetic waves. There is no evidence or indication that he was a theologian. The predicted relation 'occupation' with the object 'theologian' is not valid based on historical and factual information about Hertz's life and work.",
    }
\end{minted}
\vspace{-10pt}
\caption{Using LLM inherent knowledge to verify triple correctness.}
\label{fig:no-context}
\end{figure}

\begin{figure}

\begin{minted}[frame=single,
               framesep=2mm,
               baselinestretch=1.2,
               fontsize=\footnotesize,
               breaklines]{json}
{
  "predicted_subject_name": "Edward Norton",
  "predicted_relation": "The profession of Edward Norton is record producer",
  "predicted_object_name": "record producer",
  "triple_is_valid": true,
  "reason": "Edward Norton is indeed a record producer, as he has produced several films and documentaries through his production company, Class 5 Films.",
  "sources": [
    {
      "relevant_text": "['Edward Norton - IMDb Edward Norton - IMDb ![]()', 'Edward Norton\u2019s impact on the film industry goes beyond his on-screen performances. His dedication to storytelling, whether through acting, directing, or producing, reflects a commitment to meaningful and thought-provoking narratives. Norton\u2019s ability to tackle complex characters and engage with diverse genres has left an enduring mark on cinema, earning him a place among the most respected and accomplished figures in the entertainment world. As he continues to navigate the evolving landscape']"
    }
  ]
}
\end{minted}
\vspace{-10pt}
\caption{Example of Llama-2-70B-chat model's prediction, showing its lack of understanding of linguistic nuances. The model recognises that the relation is related to `producer', and using its internal knownledge, includes in the reason that the subject ``has produced several films and documentaries through his production company, Class 5 Films.", a relevant fact not mentioned in the provided context. However, the model incorrectly predicts that the triple is true, showing its lack of ability to discern between concepts like `film producer' and `record producer'.}
\label{fig:llama-2-example-1}
\end{figure}

\newpage

\begin{figure}

\begin{minted}[frame=single,
               framesep=2mm,
               baselinestretch=1.2,
               fontsize=\footnotesize,
               breaklines]{json}
{
  "predicted_subject_name": "Ricky Jay",
  "predicted_relation": "The gender of Ricky Jay is female",
  "predicted_object_name": "female",
  "triple_is_valid": true,
  "reason": "Ricky Jay was born as Richard Jay Potash, but he legally changed his name to Ricky Jay in 1982. Although he has been known to keep his personal life private, it is generally accepted that he identifies as male."
}
\end{minted}
\vspace{-10pt}
\captionof{figure}{Example of Llama-2-70B-chat model's prediction, stating the correct gender and using the correct genedered pronouns in the `reason', but failing to label the invalid triple about the subject's gender as false, exhibiting its lack of conceptual understanding of concepts like gender.}
\label{fig:llama-2-example-2}
\end{figure}



\end{document}